\def\year{2019}\relax
\documentclass[letterpaper]{article} 
\usepackage{aaai19}  
\usepackage{times}  
\usepackage{helvet}  
\usepackage{courier}  
\usepackage{url}  
\usepackage{graphicx}  
\usepackage{subfigure}
\usepackage{threeparttable}
\usepackage{booktabs}
\usepackage{floatrow}
\usepackage{multirow}
\usepackage{amsfonts}
\usepackage{amsmath}

\usepackage{array}
\usepackage{picins}
\usepackage{capt-of}
\usepackage{amssymb}

\begin{document}
%
\title{A Coupled Evolutionary Network for Age Estimation}

\author{Peipei Li, Yibo Hu, Ran He and Zhenan Sun \\
National Laboratory of Pattern Recognition, CASIA, Beijing, China 100190 \\
Center for Research on Intelligent Perception and Computing, CASIA, Beijing, China 100190 \\
University of Chinese Academy of Sciences, Beijing, China 100049\\
Email: \{peipei.li, yibo.hu\}@cripac.ia.ac.cn,  \{rhe, znsun\}@nlpr.ia.ac.cn}

\maketitle

\begin{abstract}
Age estimation of unknown persons is a challenging pattern analysis task due to the lacking of training data and various aging mechanisms for different people. Label distribution learning-based methods usually make distribution assumptions to simplify age estimation. However, age label distributions are often complex and difficult to be modeled in a parameter way. Inspired by the biological evolutionary mechanism, we propose a Coupled Evolutionary Network (CEN) with two concurrent evolutionary processes: evolutionary label distribution learning and evolutionary slack regression. Evolutionary network learns and refines age label distributions in an iteratively learning way. Evolutionary label distribution learning adaptively learns and constantly refines the age label distributions without making strong assumptions on the distribution patterns. To further utilize the ordered and continuous information of age labels, we accordingly propose an evolutionary slack regression to convert the discrete age label regression into the continuous age interval regression. Experimental results on Morph, ChaLearn15 and MegaAge-Asian datasets show the superiority of our method.
\end{abstract}
\section{Introduction}
\begin{figure}[t]
\setlength{\abovecaptionskip}{0cm}
\begin{center}
\includegraphics[width=1\linewidth]{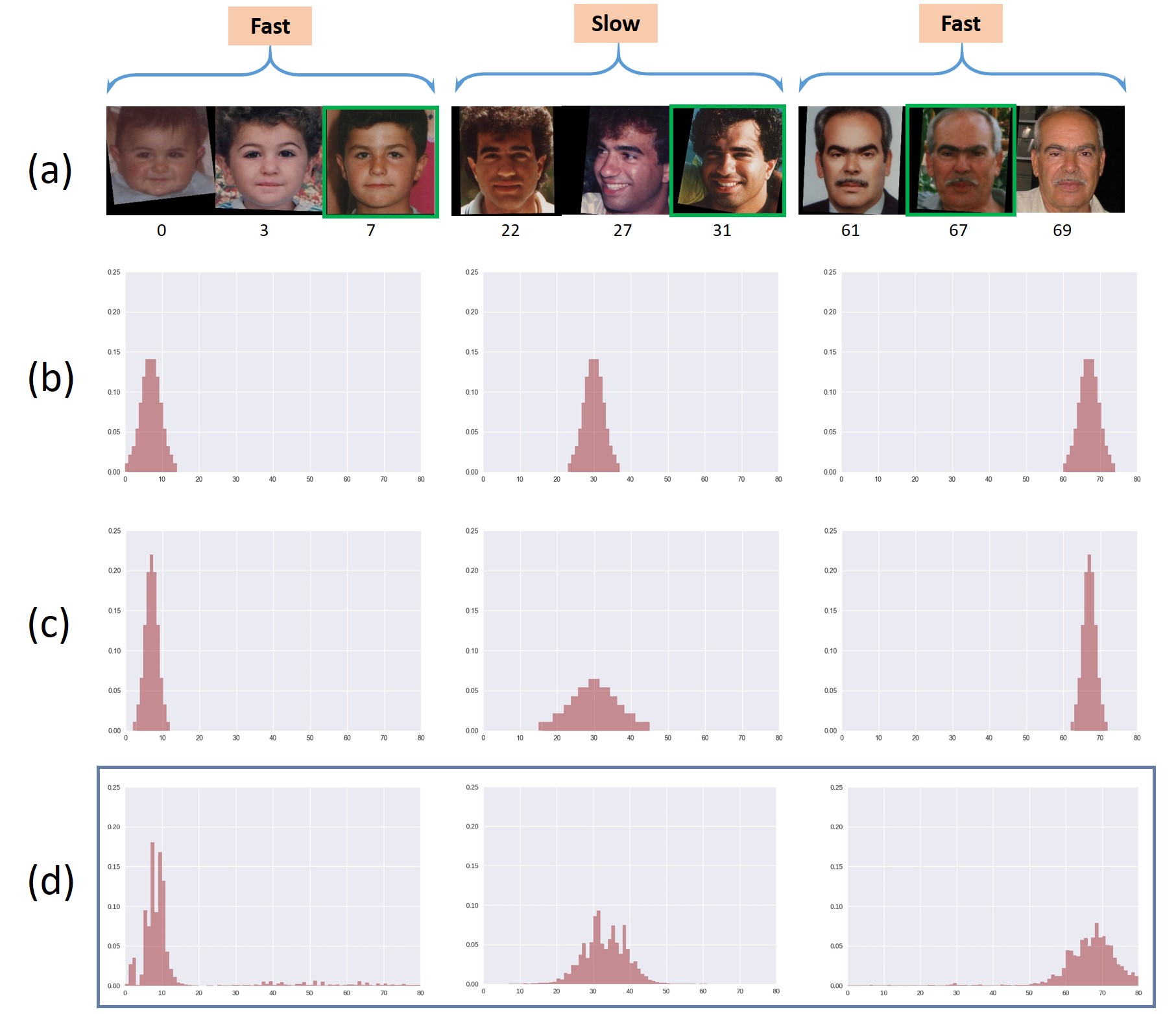}
\end{center}
\vspace{-0.2cm}
   \caption{Different label distribution assumptions for age estimation. (a) Aging speed of young-aged and old-aged are faster than middle-aged. (b) Assume that the age label distribution $X \sim N(\mu ,{\sigma ^2})$, where $\sigma$ is  same for all ages. (c) Assume that the age label distribution $X \sim N(\mu ,{\sigma ^2})$, where $\sigma$ is different at different age. (d) Learnt distribution $X$ by the proposed CEN.}
\end{figure}
Age estimation has attracted much attention in many real-world applications such as video surveillance, product recommendation, internet safety for minors, etc. It aims to label a given face image with an exact age or age group. Impressive progress has been made on age estimation in the last several decades and many methods \cite{niu2016ordinal,gao2017deep,chen2017using,shen2017deep,agustsson2017anchored,yang2018ssr} have been proposed. However, large-scale age estimation is still a very challenging problem due to several extreme reasons. 1) Many large variations with the datasets, including illumination, pose and expression, affect the accuracy of age estimation. 2) Different people age in different ways. Thus, the mapping from age-related features to age labels is not unique. 3) Age estimation is a fine-grained recognition task and it is almost impossible for human to accurately discriminate age.

Existing models for age estimation can be roughly divided into four categories: regression models \cite{shen2017deep,agustsson2017anchored}, multi-class classification models \cite{rothe2015dex,yang2018ssr}, Ranking CNN models \cite{niu2016ordinal,chen2017using} as well as label distribution learning models \cite{gao2017deep,gaoage}.  By predicting the age distribution, label distribution learning (LDL) has the potential benefits of dealing with the relevance and uncertainty among different ages. Besides, label distribution learning improves the data utilization, because the given face images provide age-related information about not only the chronological age but also its neighboring ages.


We believe that label distribution learning faces two major challenges. First, we argue that the age label distributions vary with different individuals and it is better not to assume their distribution forms like \cite{yang2015deep,gaoage}. Figure 1 depicts the detailed interpretation of this. We can see from Figure 1(a) that the aging tendencies are different for different individuals. Thus it is unreasonable to assume that the age label distributions for all ages obey Gaussian distributions with same standard deviation as Figure 1(b) shows, or with different deviations as Figure 1(c) shows. 
The second challenge is that label distribution learning is essentially a discrete learning process without considering the ordered information of age labels, while the change of age is an ordered and continuous process.

To address the first challenge, we propose evolutionary label distribution learning, a solution that uses a neural network to adaptively learn label distributions from the given individuals and constantly refine the learning results during evolution. Figure 1(d) shows the learnt distribution. It is clear that the age label distributions vary from different individuals and not strictly obey the Gaussian distribution.
For the second challenge of label distribution learning, we propose a coupled training mechanism to jointly perform label distribution learning and regression. Regression model can capture the ordered and continuous information of age labels and regress an age value, which relieves the seconde challenge. Besides, a slack term is designated to further convert the discrete age label regression to the continuous age interval regression.

The main contributions of this work are as follows:





1)	By simulating evolutionary mechanisms, we propose a Coupled Evolutionary Network (CEN) with two concurrent processes: evolutionary label distribution learning and evolutionary slack regression.

2)  The proposed evolutionary label distribution learning adaptively estimates the age distributions without the strong assumptions about the form of label distribution. Benefiting from the constant evolution of the learning results, evolutionary label distribution learning generates more precise label distributions.

3) The experiments show that the combination of label distribution learning and regression achieves superior performance. Hence, we propose evolutionary slack regression to assist evolutionary label distribution learning. Besides, we introduce a slack term to further convert the discrete age label regression to the continuous age interval regression.

4)	We evaluate the effectiveness of the proposed CEN on three age estimation benchmarks and consistently obtain the state-of-the-art results.

\section{Related Work}
\subsection{Age Estimation}
Benefiting from the deep CNNs (e.g., VGG-16 \cite{simonyan2014very}, LightCNN \cite{wu2018light}, ResNet \cite{he2016deep} and DenseNet \cite{huang2017densely}) trained on large-scale age face datasets, the deep learning based age estimation methods achieve state-of-the-art performance on age estimation, which can be roughly divided into four categories: regression \cite{shen2017deep,agustsson2017anchored}, multi-class classification \cite{rothe2015dex,can2016apparent,yang2018ssr}, Ranking CNN \cite{niu2016ordinal,chen2017using} as well as label distribution learning (LDL) \cite{gao2017deep,gaoage}.

With the huge improvement in the performance of object recognition tasks, some researchers propose to transform age estimation into a multi-classification problem, in which different ages or age groups are regarded as independent classes. However, multi-class classification methods usually neglect the relevance and uncertainty among neighboring labels. Since age is a continuous value, to better fit the aging mechanism, a natural idea is to treat age estimation as regression task. However, due to the presence of outliers, regression methods can not achieve the satisfactory results either. The change speeds of appearance at all ages are different. To alleviate this, ranking CNN and LDL methods are proposed, in which individual classifier or label distribution for each age class is adopted. In this paper, we employ LDL based method assisted with regression.

\begin{figure*}[t]
\setlength{\abovecaptionskip}{0cm}
\begin{center}
\includegraphics[width=1\linewidth]{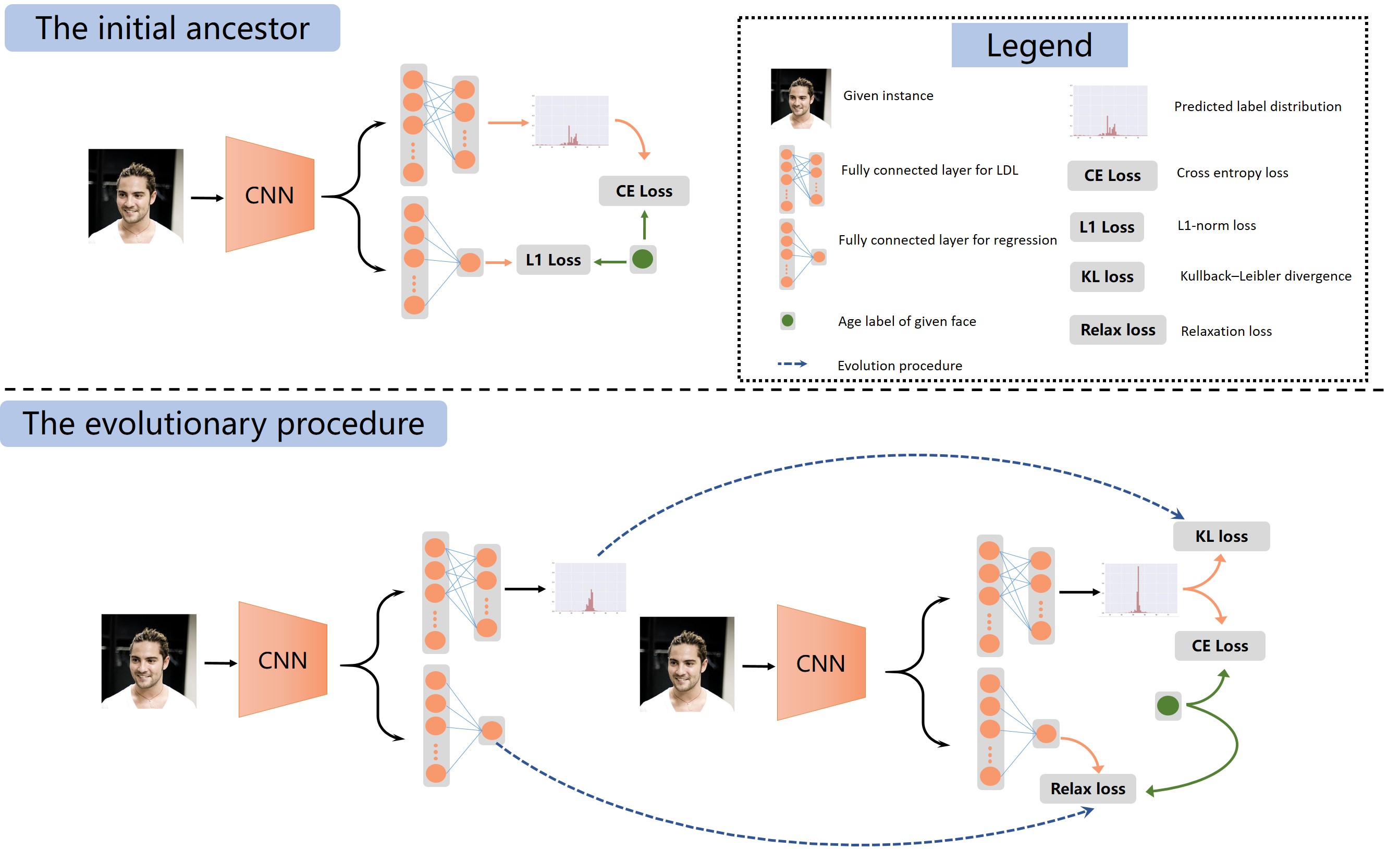}
\end{center}
   \caption{Overview of the proposed Coupled Evolutionary Network for age estimation. The initial ancestor network takes the given instance as the input and produces the initial age label distribution as well as the initial regressed age.
    The offspring network inherits the experience and knowledge of its ancestor to boost itself.}
\label{fig2}
\end{figure*}

\subsection{Label Distribution Learning}
Label ambiguity and redundancy hinder the improvement for the object recognition and classification performance. Label distribution learning (LDL) \cite{geng2013label,geng2013facial} addresses this problem by learning the distribution over each label from the description of the instance. LDL has been widely used in many applications, such as expression recognition \cite{zhou2015emotion}, public video surveillance \cite{zhang2015crowd} as well as age estimation \cite{geng2013facial,yang2016sparsity,gao2017deep,gaoage}. \cite{geng2013facial} deals with age estimation by learning the age label distribution. \cite{gaoage} analyzes that the ranking method is learning label distribution implicitly and assumes that the age label distribution is consistent with a Gaussian distribution with fixed size of standard deviation. However, since the age characteristics of different ages are different, age labels cannot be identical for all ages. To deal with it, we propose a neural network model to learn the mapping from the given image to its age label distribution.
\section{Our Approach}
In this section, we firstly give the state of problem definition. Then, we describe the two components in the proposed coupled evolutionary network (CEN). Finally, we detail the training and testing procedures, following with the network architecture.
\subsection{Problem Formulation}
In the setting of CEN, we define $L = \left[ {{l_1},{l_2}, \cdot  \cdot  \cdot ,{l_k}} \right]$ as the ages of the training set, where $l_1$ and $l_k$ are the minimal and maximum ages, respectively. Suppose
$
S = \left\{ {\left( {x},{o},{y},{l} \right)} \right\}
$
is the training set, where we omit the instance indices for simplification. Among them, \(x\) denotes the input instance and \(l \in L\) is the age of \(x\). \(o\) represents the corresponding one-hot vector of \(l\) and ${{y}}$ denotes the normalized age label, which is formulated as:
\begin{equation}
\begin{array}{c}
$${y} = \frac{{{l} - {l_1}}}{{{l_k} - {l_1}}}

$$
\end{array}
\label{e1}
\end{equation}
We are interested to learn a mapping from the instance ${x}$ to its accurate age $l$.

Inspired by the biological evolutionary mechanism, we propose a coupled evolutionary network (CEN) with two concurrent processes: evolutionary label distribution learning and evolutionary slack regression. The overall framework of CEN is depicted in Figure \ref{fig2}. We first obtain an initial ancestor CEN. Then, with the experience and knowledge transferred by the ancestor CEN, the offspring CEN utilizes and incrementally evolves itself to achieve better performance. After each evolution, the offspring CEN will be treated as the new ancestor CEN for the next evolution. The predicted age is obtained only with the last CEN.



\subsection{Evolutionary Label Distribution Learning}
Previous researches usually make strong assumptions on the form of the label distributions, which may not be able to truly and flexibly reflect the reality. We address this problem by introducing evolutionary label distribution learning, a solution that uses a neural network to adaptively learn and constantly refine the age label distributions during evolution.

The initial ancestor CEN $E_{{{\theta }_1}}$ takes the given instance ${x}$ as the input and learn to predict the age label distribution of $x$.
Then, the offspring CEN $E{_{{{\theta }_2}}}$ inherits all the age label distributions from its ancestor CEN $E{_{{{\theta }_1}}}$ and updates itself over the entire training set $S$. After each evolution, the offspring CEN $E{_{{{\bf{\theta }}_t}}}$ will be treated as the new ancestor for the next CEN $E{_{{{\bf{\theta }}_{t+1}}}}$.


\subsubsection{The Initial Ancestor}
We first utilize the initial ancestor coupled evolutionary network $E{_{{\theta _1}}}$ to adaptively learn the initial age label distributions. Specifically, given an input instance ${x}$, $E{_{{{\bf{\theta }}_1}}}$ learns the mapping from ${x}$ to the logits ${{z}^1}$ by:
\begin{equation}
\begin{array}{c}
$${{z}^1} = {\left( {{W}_{ldl}^1} \right)^{\rm T}}{{f}^1} + {{b}^{1}_{ldl}},\;\;{{z}^1} \in {R^k}$$
\end{array}
\label{e2}
\end{equation}
where ${{f}^1}$ is the output of the last pooling layer of $E{_{{{\theta }_1}}}$, ${W}_{ldl}^1$ and ${b}^{1}_{ldl}$ are the weights and biases of a fully connected layer, respectively.

The predicted age label distribution ${{p}^1} \in {R^k}$ can be formulated as:
\begin{equation}
p_i^1 = \frac{{\exp \left( {{\raise0.7ex\hbox{${z_i^1}$} \!\mathord{\left/
 {\vphantom {{z_i^1} \tau }}\right.\kern-\nulldelimiterspace}
\!\lower0.7ex\hbox{$\tau $}}} \right)}}{{\sum\nolimits_j {\exp \left( {{\raise0.7ex\hbox{${z_j^1}$} \!\mathord{\left/
 {\vphantom {{z_j^1} \tau }}\right.\kern-\nulldelimiterspace}
\!\lower0.7ex\hbox{$\tau $}}} \right)} }}
\label{e3}
\end{equation}
where $\tau$ is the temperature parameter, dominating the softness of the predicted distribution. The larger $\tau$, the softer distribution is obtained. We set $\tau = 1$ and employ cross entropy as the supervised signal to learn the initial ancestor for evolutionary label distribution learning:
\begin{equation}
\begin{array}{c}
$${L_{ce}^1 = \frac{1}{N}\sum {\left( { - \sum\limits_i {{o_i}\ln p_i^1} } \right)}} $$
\end{array}
\label{e4}
\end{equation}
where $o_i$ denotes the i-th element of the one hot vector $o$.

The goal of the initial ancestor $E{_{{\theta _1}}}$ for label distribution learning is to minimize the cross entropy loss. The predicted label distribution ${{p}^1}$ will be transferred to the offspring network $E{_{{\theta _2}}}$.

\subsubsection{The Evolutionary Procedure}
After the first evolution, we obtain the preliminary age label distribution without making strong assumptions for the form of the distribution. Then the preliminary age label distribution acts as new experience and knowledge to be transferred to the next evolution.

In $t$-th evolution, where $t > 1$, the predicted age label distribution ${{p}^t}$ of $E{_{{{\theta }_t}}}$ is calculated by Eq.(\ref{e4}). We set $\tau>1$ and employ Kullback-Leibler (KL) divergence to transfer the age label distribution from ($t$-1)-th evolution to the current evolution:
\begin{equation}
\begin{array}{c}
$$\begin{array}{l}L_{kl}^t = KL\left( {{p^{t - 1}}||{p^t}} \right) = \sum\limits_i {p_i^{t - 1}} \ln \frac{{p_i^{t - 1}}}{{p_i^t}}\\ \;\;\;\;\;\;\;\;\;\;\;= \sum\limits_i {\left( {p_i^{t - 1}\ln p_i^{t - 1} - p_i^{t - 1}\ln p_i^t} \right)} \end{array}
$$
\end{array}
\label{e5}
\end{equation}

Since ${p_i^{t - 1}\ln p_i^{t - 1}}$ is a constant, Eq.(\ref{e5}) can be further simplified as follows:
\begin{equation}
\begin{array}{c}
$$L_{kl}^t =  - \sum\limits_i {p_i^{t - 1}\ln p_i^t}
$$
\end{array}
\label{e6}
\end{equation}

It is worth nothing that there is a discrepancy between the real label distribution and the predicted label distribution ${{p}^{t-1}}$ of $E{_{{{{\theta }}_{t-1}}}}$. Using only Eq.(\ref{e6}) in the evolutionary procedure may obtain inferior performance. Consequently, we employ an additional cross entropy term ${L_{ce}^t}$ to rectify such discrepancy.

The final supervision for evolutionary procedure contains both the predicted age label distributions and the target age labels, which can be formulated as:
\begin{equation}
\begin{array}{c}
$${L_{ldl}^t} = {\alpha}L_{kl}^t + {\left(1-{\alpha}\right)}L_{ce}^t
$$
\end{array}
\label{e7}
\end{equation}
where $\alpha$ is the trade-off parameter to balance the importance of KL loss and cross entropy loss.
\subsection{Evolutionary Slack Regression}
Evolutionary label distribution learning is essentially a discrete learning process without considering the ordered information of age labels. However, the change of age is an ordered and continuous process. Accordingly, we propose a new regression method, named evolutionary slack regression, to transfer the ordered and continuous age information of the previous evolution to the current evolution. Specially, a slack term is introduced into evolutionary slack regression, which converts the discrete age label regression to the continues age interval regression.

The initial ancestor CEN $E{_{{{\bf{\theta }}_1}}}$ takes the given instance $x$ as the input and produces a roughly regressed age. Then, the absolute difference between the regressed age and the ground-truth age is treated as knowledge to be inherited by the offspring CEN $E{_{{{\bf{\theta }}_2}}}$. Similarly, after each evolution, the offspring CEN $E{_{{{\bf{\theta }}_t}}}$ will be treated as the new ancestor for the next evolution.

\subsubsection{The Initial Ancestor}
For regression, $E{_{{{\bf{\theta }}_1}}}$ learns the mapping from the given instance ${x}$ to a real value ${{s}^1} \in {R}$:
\begin{equation}
\begin{array}{c}
$${{s}^1} = {\left( {{{W}}_{reg}^1} \right)^{\rm T}}{{{f}}^1} + {{{b}}^1_{reg}}
$$
\end{array}
\label{e8}
\end{equation}
where $ {{W}_{reg}^1}$ and ${{{b}}^1_{reg}}$ are the weights and biases of a fully connected layer, respectively.

We train the initial ancestor $E{_{{{{\theta }}_1}}}$ with ${\ell _1}$ loss to minimize the distance between the regressed age ${{{s}}^1}$ and the ground-truth age $y$.
\begin{equation}
\begin{array}{c}
$$L_{\ell _1}^1 = \frac{1}{N}\sum {|{{\bf{s}}^1} - y|}
$$
\end{array}
\label{e9}
\end{equation}

\subsubsection{The Evolutionary Procedure}
We observe that the Eq.(\ref{e9}) is essentially a discrete regression process, because the target age $y$ is a discrete value. In order to deliver the ordered and continuous age information of the ancestor CEN $E{_{{{{\theta }}_{t-1}}}}$ to the offspring CEN $E{_{{{{\theta }}_t}}}$, we introduce a slack term $\Delta {{s}^{t-1}}$ into the regression of $E{_{{{{\theta }}_t}}}$, which is defined as follows:
\begin{equation}
\begin{array}{c}
$$\Delta {{{s}}^{t - 1}} = |{{{s}}^{t - 1}} - y|,\;\;t > 1
$$
\end{array}
\label{e10}
\end{equation}

We assume that $E{_{{{{\theta }}_t}}}$ is superior to $E{_{{{{\theta }}_{t-1}}}}$, which means the regression error of $E{_{{{{\theta }}_t}}}$ should not exceed $\Delta {{s}^{t-1}}$:
\begin{equation}
 - \Delta {s^{t - 1}} \le {s^t} - y \le \Delta {s^{t - 1}}
\label{e11}
\end{equation}

Eq.(\ref{e11}) can be rewritten as:
\begin{equation}
|{s^t} - y| - \Delta {s^{t - 1}} \le 0
\label{e12}
\end{equation}

Above all, we define a slack ${\ell _1}$ loss as follows:
\begin{equation}
\begin{array}{c}
$$L_{slack\_{\ell _1}}^t = \max \left( {0,|{s^t} - y| - \Delta {s^{t - 1}}} \right)

$$
\end{array}
\label{e13}
\end{equation}

Eq.(\ref{e13}) pushes the regressed age $s^t$ of $E{_{{{{\theta }}_t}}}$ lies in a continuous age interval \(\left[ {y - \Delta {s^{t - 1}},y + \Delta {s^{t - 1}}} \right]\), but not strictly equal to a discrete age label $y$. From this perspective, by introducing the slack term $\Delta {{s}^{t-1}}$ into the regression, we convert the discrete age label regression to the continuous age interval regression in age estimation.

At each evolution, we minimize the slack ${\ell _1}$ loss and find the $\Delta {{s}^{t - 1}}$ can gradually decrease.
Specially, a slack term is introduced into evolutionary slack regression, which further converts
the discrete age label regression to the continuous age interval
regression.
\subsection{Training Framework}
The training procedure of CEN contains both evolutionary label distribution learning and evolutionary slack regression. It can be divided into two parts: the initial ancestor and the evolutionary procedure.

The total supervised loss for the initial ancestor $E{_{{{\bf{\theta }}_1}}}$ is
\begin{equation}
\begin{array}{c}
$${L^1} = L_{ce}^1 + {\lambda _1}L_{\ell _1}^1
$$
\end{array}
\label{e14}
\end{equation}
where ${\lambda _1}$ is the trade-off parameter to balance the importance of the initial label distribution learning and the $\ell _1$ regression.

The total supervised loss for the evolutionary procedure is
\begin{equation}
\begin{array}{c}
$${L^t} = L_{ldl}^t + {\lambda _t}L_{slack\_{\ell _1}}^t
$$
\end{array}
\label{e15}
\end{equation}
where $t > 1$ and ${\lambda _t}$ is the trade-off parameter to balance the importance of evolutionary label distribution learning and the slack $\ell _1$ regression.
\subsection{Age Estimation in Testing}
In the testing phase, for a given instance, we use ${{\hat y}_{ldl}}$ to denote the estimated age of evolutionary label distribution learning, which can be written as:
\begin{equation}
\begin{array}{c}
$${{\hat y}_{ldl}} = \sum\limits_i {p_i^t} {l_i}
$$
\end{array}
\label{e16}
\end{equation}

The estimated age ${{\hat y}_{reg}}$ of evolutionary slack regression can be formulated as
\begin{equation}
\begin{array}{c}
$${{\hat y}_{reg}} = \left( {{l_k} - {l_1}} \right) \cdot {{s}^t} + {l_1}
$$
\end{array}
\label{e17}
\end{equation}
where ${l_1}$ and $l_k$ are the minimal and maximum ages of the training set, respectively.

Then, the final estimated age ${\hat y}$ is the average of the above two results.
\begin{equation}
\begin{array}{c}
$$\hat y = \frac{{{{\hat y}_{ldl}} + {{\hat y}_{reg}}}}{2}
$$
\end{array}
\label{e18}
\end{equation}
\subsection{Network Architecture}
\begin{figure*}[t]
\setlength{\abovecaptionskip}{0cm}
\begin{center}
\includegraphics[width=1\linewidth]{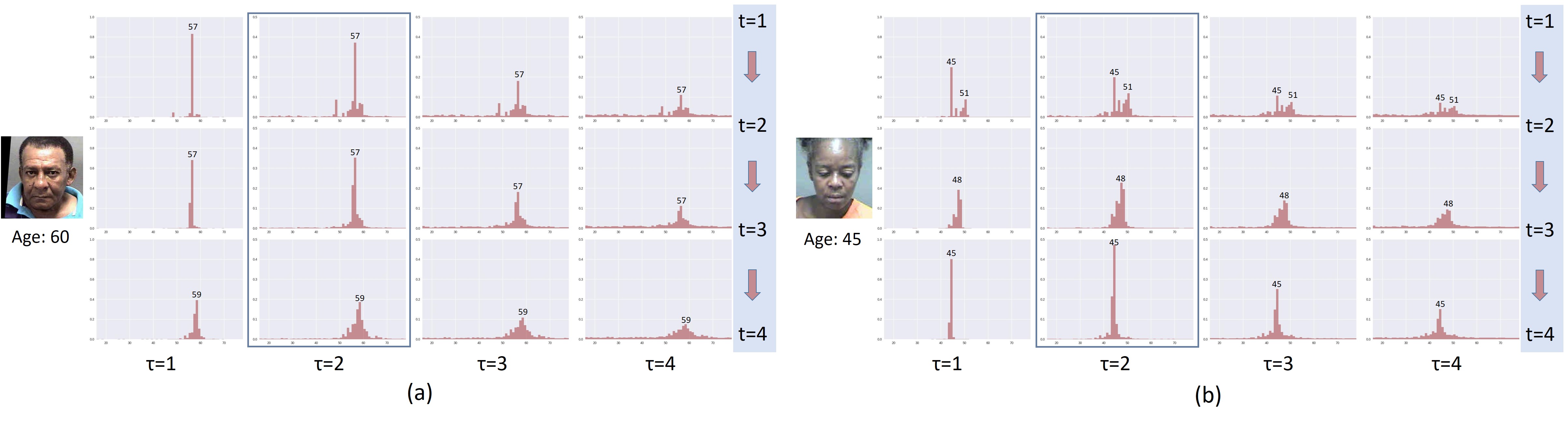}
\end{center}
\vspace{-0.2cm}
   \caption{The evolutions of the age label distributions with different temperature parameters $\tau$, where $t$ denotes the $t$-th evolution. With the given instance, the first, second and third rows are the predicted age label distributions of $E{r_{{{\bf{\theta }}_1}}}$, $E{r_{{{\bf{\theta }}_2}}}$ and $E{r_{{{\bf{\theta }}_3}}}$ respectively.}
\end{figure*}

ResNet10 and ResNet18 \cite{he2016deep} are adopted as the backbone networks of the proposed method. In particular, two fully connected layers are inserted immediately after the last pooling layer for evolutionary label distribution learning and evolutionary slack regression respectively. Considering the size and efficiency of ResNet10 and ResNet18, we further halve the number of feature channels and obtain two tiny variations, named ResNet10-Tiny and ResNet18-Tiny respectively. The details are listed in Table \ref{tab:net_architecture}.

\section{Experiments}
\subsection{Dataset and Protocol}

We evaluate the proposed CEN on both apparent age and real age datasets.

\textbf{IMDB-WIKI} \cite{rothe2015dex} is the largest publicly available dataset of facial images with age and gender labels. It consists of 523,051 facial images in total, 460,723 images from IMDB and 62,328 from Wikipedia. The ages of IMDB-WIKI dataset range from 0 to 100 years old. Although it is the largest dataset for age estimation, IMDB-WIKI is still not suitable for evaluation due to the existing of much noise. Thus, like most previous works \cite{yang2018ssr}, we utilize IMDB-WIKI only for pre-training.

\textbf{ChaLearn15} \cite{escalera2015chalearn} is the first dataset for apparent age estimation, which contains 4,691 color images, 2,476 for training, 1,136 for validation and the rest 1087 for testing. ChaLearn15 comes from the first competition track ChaLearn LAP 2015. Each image is labeled using the online voting platform. We follow the protocol in \cite{rothe2018deep} to train on the training set and evaluate on the validation set.


\textbf{Morph} \cite{ricanek2006morph} is the most popular benchmark for real age estimaion, which contains 55,134 color images of 13,617 subjects with age and gender information. The age of Morph ranges from 16 to 77 years old. It has four images of each subject on average. Classical protocol 80-20 split is used for Morph.


\textbf{MegaAge-Asian} \cite{zhang2017quantifying} is a newly released large-scale facial age dataset. Different from most of facial age datasets that only contain faces of Westerners, there are only faces of Asians in MegaAge-Asian dataset. It consists of 40, 000 images encompassing ages from 0 to 70. Following \cite{zhang2017quantifying}, we reserve 3,945 images for testing.

\subsection{Evaluation Metric}
We evaluate the performance of the proposed CEN with MAE, $\epsilon$-error and CA(n).

\textbf{Mean Absolute Error (MAE)} is widely used to evaluate the performance of age estimation. It is defined as the average of distances between the ground-truth and predicted age, which can be written as:
\begin{equation}
\begin{array}{c}
$$MAE = \frac{1}{N}\sum\limits_{i = 1}^N {|{{\hat y}_i} - {y_i}|}
$$
\end{array}
\end{equation}
where ${{\hat y}_i}$ and ${y_i}$ denote the predicted age and the ground-truth of the $i$-th testing instance, respectively.

\textbf{$\epsilon$-error} is the evaluation metric for apparent age estimation, which can be formulated as:
\begin{equation}
\begin{array}{c}
$$\epsilon- error = \frac{1}{N}\sum\limits_{i = 1}^N {\left( {1 - \exp ( - \frac{{{{\left( {{{\hat x}_i} - {\mu _i}} \right)}^2}}}{{2\sigma _i^2}})} \right)}
$$\end{array}
\end{equation}
where $ {{\hat x}_i}$, ${{\mu _i}}$ and ${{\sigma _i }}$ denote the predicted age, mean age and standard deviation of the $i$-th testing instance, respectively.

\textbf{Cumulative Accuracy (CA)} is employed as the evaluation metric for MegaAge-Asian, which can be calculated as:
\begin{equation}
\begin{array}{c}
$$CA\left( n \right) = \frac{{{K_n}}}{K} \times 100\%
$$
$$\end{array}
\end{equation}
where ${K_n}$ is the number of the test images whose absolute estimated error is smaller than $n$. We report CA(3), CA(5), CA(7) as \cite{zhang2017quantifying,yang2018ssr} in our experiments.

\subsection{Implementation Details}

\subsubsection{Pre-processing.}
We utilize multi-task cascaded CNN \cite{zhang2016joint} to detect and align face images. Then all the images are resized into 224 $\times$ 224 as the inputs. Besides, data augmentation is important to deep neural networks for age estimation. We augment the training data by: (a) random resized cropping with the aspect ratio from 0.8 to 1.25 and the scale from 0.8 to 1.0; (b) random horizontal flipping with the probability of 0.5.
\subsubsection{Training Details.}
All the network architectures used in CEN are pretrained on the IMDB-WIKI dataset by Eq.(\ref{e14}). We employ SGD optimizer and set the initial learning rate, the momentum and the weight decay to 0.01, 0.9 and 1e-4, respectively. The learning rate is decreased by a factor of 10 every 40 epochs. Each model is trained totally 160 epochs with the mini-batch size of 128. And then the pre-trained models on IMDB-WIKI are used as initializations on the target age datasets including ChaLearn15, Morph and Megaage-Asian.  All the networks are optimized by SGD and the initial learning rate, the momentum and the weight decay are set to 0.001, 0.9 and 1e-4, respectively. If not specific, we employ \({\lambda _1} = {\lambda _t} = 4\), \(\alpha  = 0.5\) and \(\tau  = 2\) in our experiments. The learning rate is decreased by a factor of 10 every 40 epochs. Each model is trained totally 160 epochs with the mini-batch size of 128.

\subsection{Analysis of Coupled Training Mechanism}
In this subsection, we explore the coupled training mechanism of label distribution learning and regression. Table \ref{tab:1} shows the comparison results.
The first and second rows are the baseline results of using only label distribution learning(LDL) and regression(Reg), respectively. The last three rows present the coupled training performance(LDL+Reg). Specifically, with coupled training mechanism, ${{\hat y}_{ldl}}$, ${{\hat y}_{reg}}$ and ${\hat y}$ are calculated by Eq.(\ref{e16}),Eq.(\ref{e17}) and Eq.(\ref{e18}), respectively, denoting the outputs of label distribution learning, regression and average of the above outputs.

\begin{table}[htbp]\scriptsize
\centering
\caption{Comparisons with using only label distribution learning and regression on Morph and MegaAge-Asian. Lower MAE is better, while higher CA(n) is better. We employ ResNet-18 as the backbone. The unit of CA(n) is $\%$.}

\begin{tabular}{l|l|llllcl}
\hline
\multirow{2}{*}{Methods} & \multicolumn{1}{c|}{ Morph} & \multicolumn{3}{c}{MegaAge-Asian}\\
\cline{2-5}

&MAE&CA(3)&CA(5)&CA(7)\\
\hline
\hline
Reg &2.578&58.22 &79.01&89.03\\
LDL &2.323& 59.14&78.70&89.26\\
\hline
LDL+Reg $\left({{\hat y}_{reg}}\right)$& 2.243 &60.57 &79.77&90.21\\
LDL+Reg $\left({{\hat y}_{ldl}}\right)$& 2.231 &59.14 &79.24&89.62\\
LDL+Reg $\left({\hat y}\right)$& \textbf{2.220} &\textbf{60.83} &\textbf{80.11}&\textbf{90.52}\\
\hline
\end{tabular}\label{tab:1}
\end{table}

Obviously, the proposed coupled training mechanism (LDL+Reg) achieves superior performance than training only with LDL or Reg. For example, compared with Reg, LDL+Reg ${{\hat y}_{reg}}$ gains 0.335 improvement of MAE on Morph. And the average of the label distribution learning and regression terms ${\hat y}$ further gains 0.023 and 0.011 improvements of MAE compared with ${{\hat y}_{reg}}$ and ${{\hat y}_{ldl}}$, respectively. It indicates that the coupled training mechanism can significantly improve the performance of age estimation task, therefore we choose ${\hat y}$ as age estimation results in the following experiments.

\subsection{Comparisons with State-of-the-Arts}
We compare the proposed CEN with previous state-of-the-art methods on Morph, ChaLearn and MegaAge-Asian datasets. The proposed CEN performs mostly the best among all the state-of-the-art methods.

Table \ref{tab:morph} shows the MAEs of the individual methods on Morph. Benefiting from the adaptive learning of label distribution and the coupled evolutionary mechanism, our CEN, based on ResNet-18, obtains \textbf{1.905} on Morph and outperforms the previous state-of-the-art method from ThinAgeNet \cite{gaoage}.

\begin{table}[htbp]\scriptsize
\centering
\caption{Comparisons with state-of-the-art methods on the Morph dataset. Lower MAE is better.}
\begin{threeparttable}
\begin{tabular}{ll|llclclcc}
\hline
\multirow{2}{*}{Methods} & \multirow{2}{*}{Pretrained} & Morph \\
\cline{3-3}
&&MAE\\
\hline
\hline
OR-CNN\cite{niu2016ordinal} & - & 3.34 \\
DEX\cite{rothe2018deep} & IMDB-WIKI$^ *$ & 2.68\\
Ranking \cite{chen2017using} &Audience& 2.96 \\
Posterior\cite{zhang2017quantifying} & IMDB-WIKI &2.52\\
DRFs\cite{shen2017deep}&- &2.17\\
SSR-Net\cite{yang2018ssr}&IMDB-WIKI &2.52\\
M-V Loss\cite{pan2018mean}&IMDB-WIKI &2.16\\
TinyAgeNet \cite{gaoage} &MS-Celeb-1M$^ *$ & 2.291\\
ThinAgeNet \cite{gaoage} &MS-Celeb-1M$^ *$ & 1.969\\

\hline

CEN(ResNet10-Tiny)& IMDB-WIKI&2.229\\
CEN(ResNet10)& IMDB-WIKI&2.134\\
CEN(ResNet18-Tiny)& IMDB-WIKI&2.069\\
CEN(ResNet18)& IMDB-WIKI&\textbf{1.905}\\

\hline
\end{tabular}\label{tab:morph}
\begin{tablenotes}
        \footnotesize
        \item[*] \tiny{Used partial data of the dataset;}
      \end{tablenotes}
\end{threeparttable}
\end{table}

In addition to real age estimation, apparent age estimation is also important. We conduct experiments on ChaLearn15 to validate the performance of our method on apparent age estimation. Since there are only 2,476 training data in ChaLearn15, huge network may lead to overfitting. Therefore, we choose ResNet10-Tiny with 1.2M parameters as the backbone for evaluations.  Table \ref{tab:cha} shows the comparison results of MAE and ${\epsilon}$-error. The proposed method creates a new state-of-the-art 3.052 of MAE. The ${\epsilon}$-error 0.274 is also close to the best competition result 0.272 (ThinAgeNet). Note that the parameters of CEN(ResNet10-Tiny) is 1.2M, less than 3.7M of ThinAgeNet.
\begin{table}[htbp]\scriptsize
\centering
\caption{Comparisons with state-of-the-art methods on the ChaLearn15 dataset. Lower MAE and ${\epsilon}$-error are better.}
\begin{threeparttable}
\begin{tabular}{ll|ll|llccclcl}
\hline
\multirow{2}{*}{Methods} & \multirow{2}{*}{Pretrained}  & \multicolumn{2}{c|}{ChaLearn15}&\multirow{2}{*}{ $\#$Param}\\
\cline{3-4}
&&MAE&${\epsilon}$-error&\\
\hline
\hline
DEX\cite{rothe2018deep} & -  &5.369 &0.456&134.6M\\
DEX\cite{rothe2018deep} &  \tiny{IMDB-WIKI$^ *$} &3.252&0.282&134.6M \\
ARN (Agustsson et al. 2017)&  \tiny{IMDB-WIKI}&3.153&$ $-&134.6M\\
TinyAgeNet \cite{gaoage} &\tiny{MS-Celeb-1M$^ *$}  & 3.427&0.301&\textbf{0.9M} \\
ThinAgeNet \cite{gaoage} &\tiny{MS-Celeb-1M$^ *$}  & 3.135&\textbf{0.272}&3.7M\\
\hline

CEN(ResNet10-Tiny)& \tiny{IMDB-WIKI}&\textbf{3.052}&0.274&1.2M\\
\hline
\end{tabular}\label{tab:cha}
\begin{tablenotes}
        \footnotesize
        \item[*] \tiny{Used partial data of the dataset;}
      \end{tablenotes}
\end{threeparttable}
\end{table}

Besides, we evaluate the performance of CEN on the MegaAge-Asian dataset, which only contains Asians. Table \ref{tab:asian} reports the comparison results of CA(3), CA(5) and CA(7). Our CEN(ResNet18-Tiny) achieves 64.23\%, 82.15\% and 90.80\%, which are the new state-of-the-arts, and obtains 0.22\%, 0.80\% and 1.18\% improvements compared with previous best method Posterior\cite{zhang2017quantifying}.
\begin{table}[htbp]\scriptsize
\centering
\caption{Comparisons with state-of-the-art methods on the MegaAge-Asian dataset. The unit of CA(n) is $\%$. Higher CA(n) is better.}
\begin{threeparttable}
\begin{tabular}{ll|lllc}
\hline
\multirow{2}{*}{Methods} & \multirow{2}{*}{Pretrained} & \multicolumn{3}{c}{MegaAge-Asian}\\
\cline{3-5}
&&CA(3)&CA(5)&CA(7)\\
\hline
\hline
Posterior\cite{zhang2017quantifying} & IMDB-WIKI&  62.08& 80.43&90.42\\
Posterior\cite{zhang2017quantifying} & MS-Celeb-1M&  64.23& 82.15&90.80\\
MobileNet\cite{yang2018ssr}& IMDB-WIKI&44.0&60.6&-\\
DenseNet\cite{yang2018ssr}& IMDB-WIKI&51.7&69.4&-\\
SSR-Net\cite{yang2018ssr}& IMDB-WIKI&54.9&74.1&-\\
\hline

CEN(ResNet10-Tiny)& IMDB-WIKI&63.60&82.36&91.80\\
CEN(ResNet10)& IMDB-WIKI&62.86&81.47&91.34\\
CEN(ResNet18-Tiny)& IMDB-WIKI&\textbf{64.45}&\textbf{82.95}&\textbf{91.98}\\
CEN(ResNet18)& IMDB-WIKI&63.73&82.88&91.64\\

\hline
\end{tabular}\label{tab:asian}
\end{threeparttable}
\end{table}

\subsection{The Superiority of Evolutionary Mechanism}

In this subsection, we qualitatively and quantitatively demonstrate the superiority of the proposed evolutionary mechanism. Figure 3 depicts the evolutions of age label distributions. As shown in the second column of Figure 3(b), with the given instance who is 45 years old, the first predicted distribution can be approximately regarded as a bimodal distribution with two peaks 41 and 51, which is ambiguous for age estimation. After 1 time evolution, the predicted distribution is refined from bimodal distribution to unimodal distribution with the single peak 48. After 2 times evolution, the peak of unimodal distribution moves from 48 to 45, which is the true age of the input instance. This movement indicates the effectiveness of the additional cross entropy term in Eq.(\ref{e7}), which aims to rectify the discrepancy between the real label distribution and the predicted label distribution. More results are shown in Figure 4 and Figure 5.

\begin{table}[htbp]\scriptsize
\centering
\caption{The influences of evolution mechanism. The first evolution($t=1$) means the initial ancestor in CEN. The unit of CA(n) is $\%$. Lower MAE is better, while higher CA(n) is better.}
\begin{threeparttable}
\begin{tabular}{l|l|l|llllcl}
\hline
\multicolumn{2}{c|}{\multirow{2}{*}{Backbones}} &  \multicolumn{1}{c|}{Morph} & \multicolumn{3}{c}{MegaAge-Asian}\\
\cline{3-6}
\multicolumn{2}{c|}{}&MAE&CA(3)&CA(5)&CA(7)\\
\hline
\hline
\multirow{4}{*}{CEN(ResNet10-Tiny)}&t=1&2.446&60.52 &80.13&90.64\\
&t=2&2.300  & 62.01&81.90&91.64\\
&t=3&2.241  & 63.14&82.31&\textbf{91.84}\\
&t=4&\textbf{2.229}  &\textbf{63.60} &\textbf{82.36}&91.80\\
\hline

\multirow{4}{*}{CEN(ResNet10)}&t=1&2.321  &59.57 &79.44&89.39\\
&t=2&2.207  &61.91 &81.18&91.16\\
&t=3&2.150  &\textbf{62.86} &81.47&\textbf{91.34}\\
 &t=4&\textbf{2.134}  &62.78 &\textbf{81.77}&91.00\\
\hline

\multirow{4}{*}{CEN(ResNet18-Tiny)}&t=1&2.304  &61.88 &81.31&91.34\\
&t=2&2.136  & 63.57&82.00&91.46\\
&t=3&\textbf{2.069}  &\textbf{64.52} &82.03&91.70\\
&t=4&2.074  &64.45 &\textbf{82.95}&\textbf{91.98}\\
\hline

\multirow{4}{*}{CEN(ResNet18)}&t=1&2.220  &60.83 &80.11&90.52\\
&t=2&1.996  &62.42 &82.75&91.59\\
&t=3&\textbf{1.905}  &63.31 &\textbf{83.11}&\textbf{92.28}\\
&t=4&1.919  &\textbf{63.73} &82.88&91.64\\

\hline
\end{tabular}\label{tab:evolutionary}
\end{threeparttable}

\end{table}

In addition, we show quantitative experimental results of evolutionary mechanism on Morph and MegaAge-Asian in Table \ref{tab:evolutionary}. We observe that the performance of all the network architectures will increase through evolution. For example, after 2 time evolutions (from $t=1$ to $t=3$), the CA(7) for CEN(ResNet10-Tiny), CEN(ResNet10), CEN(ResNet18-Tiny) and CEN(ResNet18) on MegaAge-Asian improve from 90.64\%, 89.39\%, 91.34\% and 90.52\% to 91.84\%, 91.34\%, 91.70\% and 92.28\%, respectively. It demonstrates the superiority of the proposed evolutionary mechanism. Specifically, there is a significant improvement from the first evolution($t=1$) to the second evolution($t=2$), which is mainly because of the additional employment of Kullback-Leibler (KL) divergence and the slack term. We also observe that the best results are achieved in 3-th evolution or 4-th evolution, indicating the boosting is saturated in the evolutionary procedure.

Additional visualization results of the evolutionary age label distributions on Morph and MegaAge-Asian are presented in Figure 4 and Figure 5.
\subsection{Ablation Study}
In this section, we explore the influences of three hyper-parameters $\tau$, $\alpha$ and $\lambda$ for CEN. All the ablation studies are trained on Morph with ResNet18 model.
\subsubsection{Influence of Temperature Parameters $\tau$.}

The temperature parameter $\tau$ plays an important role in the age distribution estimation. Figure 3 provides a schematic illustration of the influence of $\tau$. In Figure 3(a), from left to right, each column presents the age label distributions when $\tau=1,2,3,4$. We observe that $\tau=2$ works better in our CEN than other lower or higher temperatures. To be specific, when $\tau=1$, the negative logits are mostly ignored, even though they may convey useful information about the knowledge from the ancestor CEN. While $\tau=3$ or $4$ would suppress the probability of peak in the age label distribution, which contributes to misleading during optimization.

Besides, we quantitatively compare the MAE on Morph with different $\tau$. Specifically, we fix $\alpha$ to 0.5, $\lambda$ to 2 and report results with $\tau$ ranging from 1 to 5 in Table \ref{tab:tem}. Apparently, when $\tau=2$, we obtain the best result on MAE 1.905. Thus, we choose to use $\tau=2$ in our experiments.
\begin{table}[htbp]\scriptsize
\centering
\caption{The influences of hyper-parameters $\lambda$, $\alpha$ and $\tau$. }

\begin{tabular}{p{0.15cm}p{0.15cm}p{0.15cm}|c|p{0.15cm}p{0.15cm}p{0.15cm}|l|p{0.15cm}p{0.15cm}p{0.15cm}|c}
\hline
\multicolumn{3}{c|}{Hyper-param} &  Morph&\multicolumn{3}{c|}{Hyper-param} &  Morph&\multicolumn{3}{c|}{Hyper-param} &  Morph \\
\hline
$\tau$&\multicolumn{1}{c}{$\alpha$}&$\lambda$&MAE&$\tau$&\multicolumn{1}{c}{$\alpha$}&$\lambda$&MAE&$\tau$&\multicolumn{1}{c}{$\alpha$}&$\lambda$&MAE\\
\hline
\hline
1&0.5&4 &2.096 &2&0.25&4 &1.946&2&0.5&1 &1.965\\
\textbf{2}&0.5&4   &\textbf{1.905} &2&\textbf{0.50}&4 &\textbf{1.905}&2  &0.5&2&1.962\\
3&0.5&4 &1.941&2&0.75& 4 &1.921&2&0.5&3 &1.922\\
4&0.5&4 &1.970&2&1.00&4 &1.952&2&0.5&\textbf{4} &\textbf{1.905}\\
-&-&-&-&-&-&-&-&2&0.5&5 &1.933\\
\hline
\end{tabular}\label{tab:tem}
\end{table}

\subsubsection{Influence of Hyper-parameters $\alpha$.}
We use the hyper-parameter $\alpha$ to balance the importance of the cross entropy and Kullback-Leibler (KL) divergence losses in evolutionary label distribution learning. We fix the $\tau$ to 2, $\lambda$ to 2 and report results with $\alpha$ from 0.25 to 1.00 in Table \ref{tab:tem}. When $\alpha=0.50$, we obtain the best result, which indicates that both the cross entropy loss and Kullback-Leibler divergence loss are equally important ($\alpha=0.50$) in our method.
\subsubsection{Influence of Hyper-parameters $\lambda$.}
We use the hyper-parameter $\lambda$ to balance the importance of the evolutionary label distribution learning and evolutionary slack regression in the our CEN. We fix the $\tau$ to 2, $\alpha$ to 0.5 and report results with $\lambda$ from 1 to 4 in Table \ref{tab:tem}. We can see that when $\lambda=4$, CEN performs the best.

\section{Conclusion}
In this paper, we propose a Coupled Evolutionary Network (CEN) for age estimation, which contains two concurrent processes: evolutionary label distribution learning and evolutionary slack regression. The former contributes to adaptively learn and refines the age label distributions without making strong assumptions about
the distribution patterns in an evolutionary manner. The later concentrates on the ordered and continuous information of age labels, converting the discrete age label regression to the continuous age interval regression. Experimental results on Morph, ChaLearn15 and MegaAge-Asian datasets show the superiority of CEN.

{
\bibliographystyle{aaai}
\bibliography{egbib}
}
\begin{table*}
\centering
\caption{Network architectures in our method.}
\vspace{0.2cm}
\renewcommand\arraystretch{1.5}
\begin{tabular}{c|c|c|c|c|c}
\hline
Layer Name & Output Size & ResNet10 & ResNet18 & ResNet10-Tiny & ResNet18-Tiny \\
\hline
Conv1 & 112 $\times$ 112 & \multicolumn{2}{c|}{7 $\times$ 7, 64, Stride 2} & \multicolumn{2}{c}{7 $\times$ 7, 32, Stride 2} \\
\hline
\multirow{2}{*}{Conv2$\_$x} & \multirow{2}{*}{56 $\times$ 56} & \multicolumn{4}{c}{3 $\times$ 3 max pool, Stride 2} \\
\cline{3-6}
& &
\({\left[ {\begin{array}{*{20}{c}}
{3 \times 3,64}\\
{3 \times 3,64}
\end{array}} \right] }\times 1\) &
\({\left[ {\begin{array}{*{20}{c}}
{3 \times 3,64}\\
{3 \times 3,64}
\end{array}} \right]} \times 2\) &
\({\left[ {\begin{array}{*{20}{c}}
{3 \times 3,32}\\
{3 \times 3,32}
\end{array}} \right]} \times 1\) &
\(\left[ {\begin{array}{*{20}{c}}
{3 \times 3,32}\\
{3 \times 3,32}
\end{array}} \right] \times 2\) \\
\hline
Conv3$\_$x & 28 $\times$ 28 &
\({\left[ {\begin{array}{*{20}{c}}
{3 \times 3,128}\\
{3 \times 3,128}
\end{array}} \right]} \times 1\) &
\(\left[ {\begin{array}{*{20}{c}}
{3 \times 3,128}\\
{3 \times 3,128}
\end{array}} \right] \times 2\) &
\(\left[ {\begin{array}{*{20}{c}}
{3 \times 3,64}\\
{3 \times 3,64}
\end{array}} \right] \times 1\) &
\(\left[ {\begin{array}{*{20}{c}}
{3 \times 3,64}\\
{3 \times 3,64}
\end{array}} \right] \times 2\) \\
\hline
Conv4$\_$x & 14 $\times$ 14 &
\(\left[ {\begin{array}{*{20}{c}}
{3 \times 3,256}\\
{3 \times 3,256}
\end{array}} \right] \times 1\) &
\({\left[ {\begin{array}{*{20}{c}}
{3 \times 3,256}\\
{3 \times 3,256}
\end{array}} \right]} \times 2\) &
\(\left[ {\begin{array}{*{20}{c}}
{3 \times 3,128}\\
{3 \times 3,128}
\end{array}} \right] \times 1\) &
\(\left[ {\begin{array}{*{20}{c}}
{3 \times 3,128}\\
{3 \times 3,128}
\end{array}} \right] \times 2\) \\
\hline
Conv5$\_$x &  7 $\times$  7 &
\(\left[ {\begin{array}{*{20}{c}}
{3 \times 3,512}\\
{3 \times 3,512}
\end{array}} \right] \times 1\) &
\(\left[ {\begin{array}{*{20}{c}}
{3 \times 3,512}\\
{3 \times 3,512}
\end{array}} \right] \times 2\) &
\(\left[ {\begin{array}{*{20}{c}}
{3 \times 3,256}\\
{3 \times 3,256}
\end{array}} \right] \times 1\) &
\(\left[ {\begin{array}{*{20}{c}}
{3 \times 3,256}\\
{3 \times 3,256}
\end{array}} \right] \times 2\) \\
\hline
 & 1 $\times$ 1 & \multicolumn{4}{c}{average pool, num$\_$age-d fc, 1-d fc} \\
\hline
\multicolumn{2}{c|}{$\#$Param} & 4.9M & 11.2M & 1.2M & 2.8M \\
\hline

\end{tabular}\label{tab:net_architecture}
\end{table*}
\newpage
\begin{figure*}
\setlength{\abovecaptionskip}{0cm}
\begin{center}
\includegraphics[width=1\linewidth]{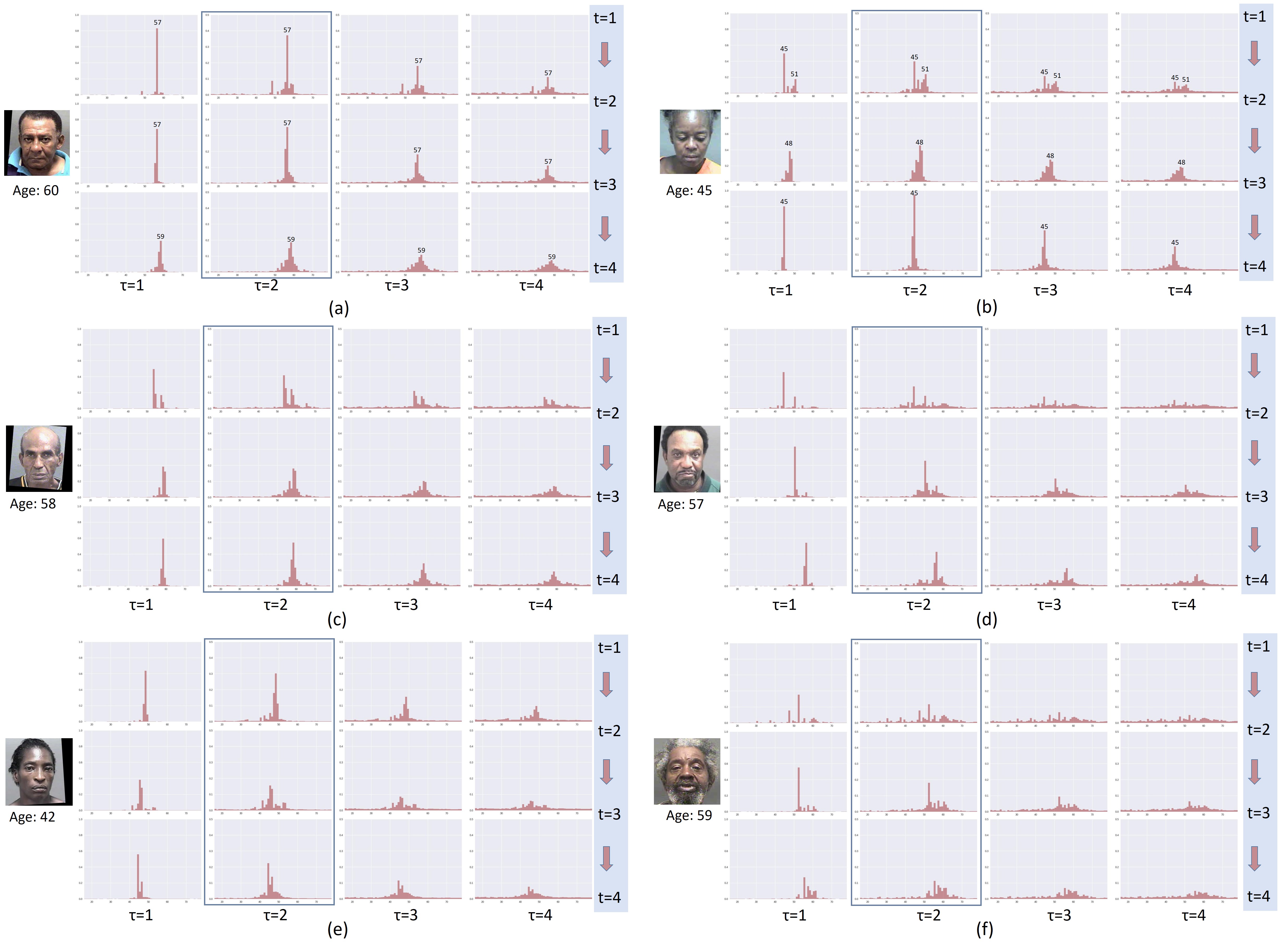}
\end{center}
\vspace{-0.2cm}
   \caption{The evolution of the age label distributions with different temperature parameters $\tau$ on Morph.}
\end{figure*} \label{fig:ad11}

\begin{figure*}
\setlength{\abovecaptionskip}{0cm}
\begin{center}
\includegraphics[width=1\linewidth]{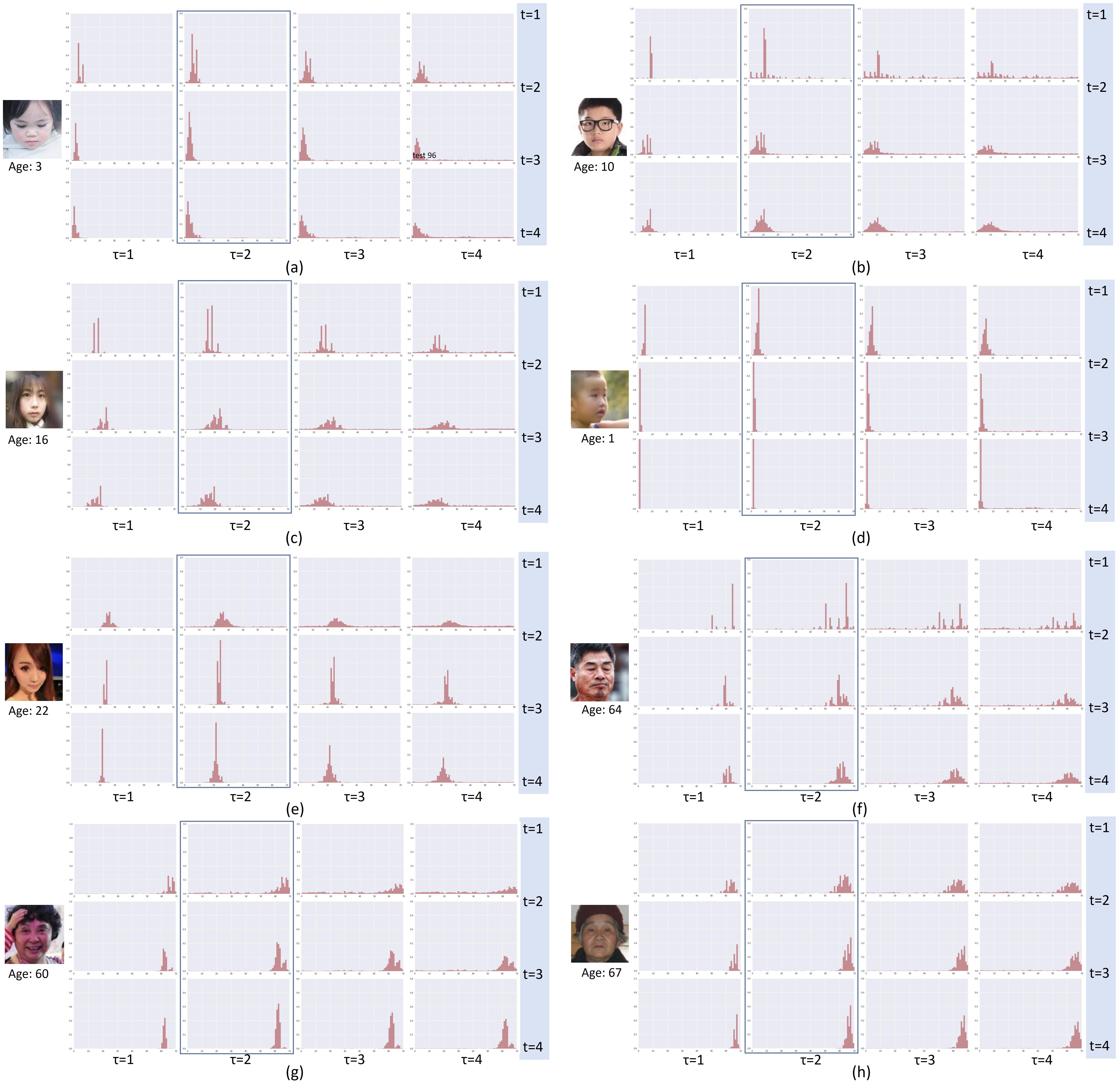}
\end{center}
\vspace{-0.2cm}
   \caption{The evolution of the age label distributions with different temperature parameters $\tau$ on MegaAge-Asian.}
\end{figure*} \label{fig:ad12}

\end{document}